\documentclass[graybox]{archivesofdatascience}
\usepackage[T1]{fontenc}
\usepackage{mathptmx}       
\usepackage{helvet}         
\usepackage{courier}        
\usepackage{type1cm}        

\usepackage{graphicx}        
\usepackage{multicol}        
\usepackage[bottom]{footmisc}

\usepackage{amsmath}		
\usepackage{amsfonts}		
\usepackage{latexsym}		
\usepackage{natbib}			

\usepackage{booktabs}		
\usepackage{placeins}		
\usepackage{url}			
\usepackage{hyperref}		

\usepackage{multirow}       
\usepackage{footnote}

\usepackage{xspace}

\usepackage{todonotes}

\usepackage{array}
\newcolumntype{L}[1]{>{\raggedright\let\newline\\\arraybackslash\hspace{0pt}}m{#1}}
\newcolumntype{C}[1]{>{\centering\let\newline\\\arraybackslash\hspace{0pt}}m{#1}}
\newcolumntype{R}[1]{>{\raggedleft\let\newline\\\arraybackslash\hspace{0pt}}m{#1}}

\newcommand{\ktvplain}{KG2Vec\xspace}
\newcommand{\ktv}{\textit{\ktvplain}\xspace}
\newcommand{\wtv}{\textit{Word2Vec}\xspace}
\newcommand{\dtv}{\textit{Doc2Vec}\xspace}
\newcommand{\rtv}{\textit{RDF2Vec}\xspace}
\newcommand{\kgl}{\textit{KGloVe}\xspace}

\makesavenoteenv{tabular}
\makesavenoteenv{table}

\Journal{Archives of Data Science, Series A\\ (Online First)}
\Volume{-}
\Number{-}
\Year{-}
\Publisher{KIT Scientific Publishing}
\ISSN{2363-9881}
\DOI{10.5445/KSP/XXXXXXXX/XX}
\newcommand{\creativecommons}{\includegraphics[height=6pt]{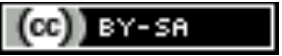}}
\usepackage[activate={true,nocompatibility}, final, tracking=true, kerning=true, spacing=true, factor=1100, stretch=10, shrink=10]{microtype}

\begin{document}


\title*{Expeditious Generation of Knowledge Graph Embeddings}

\author{Tommaso Soru, Stefano Ruberto, Diego Moussallem, Andr\'e Valdestilhas, Alexander Bigerl, Edgard Marx and Diego Esteves}
\authorrunning{Tommaso Soru et al.}
\institute{Tommaso Soru, Diego Moussallem, Andr\'e Valdestilhas, Edgard Marx \at AKSW, University of Leipzig, Germany \email{tsoru,moussallem,valdestilhas,marx@informatik.uni-leipzig.de}
\and Stefano Ruberto \at University of Pennsylvania, United States
\email{stefano.ruberto@pennmedicine.upenn.edu}
\and Alexander Bigerl \at DICE, Paderborn University, Germany
\email{alexander.bigerl@uni-paderborn.de}
\and Diego Esteves \at SDA, University of Bonn, Germany
\email{esteves@cs.uni-bonn.de}
}

\maketitle

\index{SORU, T@\emph{SORU, T}}
\index{RUBERTO, S@\emph{RUBERTO, S}}
\index{MOUSSALLEM, D@\emph{MOUSSALLEM, D}}
\index{VALDESTILHAS, A@\emph{VALDESTILHAS, A}}
\index{BIGERL, A@\emph{BIGERL, A}}
\index{MARX, E@\emph{MARX, E}}
\index{ESTEVES, D@\emph{ESTEVES, D}}
\index{VECTOR SPACE MODELS}
\index{REPRESENTATION LEARNING}
\index{KNOWLEDGE GRAPH EMBEDDING}
\index{LINK PREDICTION}

\begin{svgraybox}
This work was accepted for presentation at the 5th European Conference on Data Analysis (ECDA 2018) under the title ``A Simple and Fast Approach to Knowledge Graph Embedding''.
\end{svgraybox}

\abstract{
Knowledge Graph Embedding methods aim at representing entities and relations in a knowledge base as points or vectors in a continuous vector space.
Several approaches using embeddings have shown promising results on tasks such as link prediction, entity recommendation, question answering, and triplet classification.
However, only a few methods can compute low-dimensional embeddings of very large knowledge bases without needing state-of-the-art computational resources.
In this paper, we propose \ktv, a simple and fast approach to Knowledge Graph Embedding based on the skip-gram model.
Instead of using a predefined scoring function, we learn it relying on Long Short-Term Memories.
We show that our embeddings achieve results comparable with the most scalable approaches on knowledge graph completion as well as on a new metric.
Yet, \ktv can embed large graphs in lesser time by processing more than 250 million triples in less than 7 hours on common hardware.
}

\section{Introduction}

Recently, the number of public datasets in the Linked Data cloud has significantly grown to almost 10 thousands.
At the time of writing, at least four of these datasets contain more than one billion triples each.\footnote{\url{http://lodstats.aksw.org}}
This huge amount of available data has become a fertile ground for Machine Learning and Data Mining algorithms.
Today, applications of machine-learning techniques comprise a broad variety of research areas related to Linked Data, such as Link Discovery, Named Entity Recognition, and Structured Question Answering. 
The field of Knowledge Graph Embedding (KGE) has emerged in the Machine Learning community during the last five years.
The underlying concept of KGE is that in a knowledge base, each entity and relation can be regarded as a vector in a continuous space.
The generated vector representations can be used by algorithms employing machine learning, deep learning, or statistical relational learning to accomplish a given task.
Several KGE approaches have already shown promising results on tasks such as link prediction, entity recommendation, question answering, and triplet classification~\citep{TransG/xiao2015transg,PTransE/DBLP:journals/corr/LinLS15,TransR/lin2015learning,nickel2016holographic}. 
Moreover, Distributional Semantics techniques (e.g., \wtv or \dtv) are relatively new in the Semantic Web community.
The \rtv approaches~\citep{ristoski2016rdf2vec,cochez2017biased} are examples of pioneering research and to date, they represent the only option for learning embeddings on a large knowledge graph without the need for state-of-the-art hardware.
To this end, we devise the \ktv approach, which comprises skip-gram techniques for creating embeddings on large knowledge graphs in a feasible time but still maintaining the quality of state-of-the-art embeddings.
Our evaluation shows that \ktv achieves a vector quality comparable to the most scalable approaches and can process more than 250 million triples in less than 7 hours on a machine with suboptimal performances.


\section{Related Work} \label{sec:related}

An early effort to automatically generate features from structured knowledge was proposed in~\citep{cheng2011automated}.
RESCAL~\citep{nickel2011three} is a relational-learning algorithm based on Tensor Factorization using Alternating Least-Squares which has showed to scale to large RDF datasets such as YAGO\citep{nickel2012factorizing} and reach good results in the tasks of link prediction, entity resolution, or collective classification~\citep{nickel2014reducing}.
Manifold approaches which rely on translations have been implemented so far~\citep{TransE/bordes2013translating,TransH/wang2014knowledge,TransA/jia2015locally,TransR/lin2015learning,TransRrules/wang2015knowledge,TransG/xiao2015transg}. 
TransE is the first method where relationships are interpreted as translations operating on the low-dimensional embeddings of the entities~\citep{TransE/bordes2013translating}.
On the other hand, TransH models a relation as a hyperplane together with a translation operation on it~\citep{TransH/wang2014knowledge}.
TransA explores embedding methods for entities and relations belonging to two different knowledge graphs finding the optimal loss function~\citep{TransA/jia2015locally}, whilst PTransE relies on paths to build the final vectors~\citep{PTransE/DBLP:journals/corr/LinLS15}.
The algorithms TransR and CTransR proposed in \cite{TransR/lin2015learning} aim at building entity and relation embeddings in separate entity space and relation spaces, so as to learn embeddings through projected translations in the relation space;
an extension of this algorithm makes use of rules to learn embeddings~\citep{TransRrules/wang2015knowledge}.
An effort to jointly embed structured and unstructured data (such as text) was proposed in \cite{KGtext/wang2014knowledge}.
The idea behind the DistMult approach is to consider entities as low-dimensional vectors learned from a neural network and relations as bilinear and/or linear mapping functions~\cite{yang2014embedding}.
TransG, a generative model address the issue of multiple relation semantics of a relation, has showed to go beyond state-of-the-art results~\citep{TransG/xiao2015transg}.
ComplEx is based on latent factorization and, with the use of complex-valued embeddings, it facilitates composition and handles a large variety of binary relations~\cite{trouillon2016complex}.
The fastText algorithm was meant for word embeddings, however \cite{joulin2017fast} showed that a simple bag-of-words can generate surprisingly good KGEs.

The field of KGE has considerably grown during the last two years, earning a spot also in the Semantic Web community. 
In 2016, \cite{nickel2016holographic} proposed HolE, which relies on holographic models of associative memory by employing circular correlation to create compositional representations. 
HolE can capture rich interactions by using correlation as the compositional operator but it simultaneously remains efficient to compute, easy to train, and scalable to large datasets. 
In the same year, \cite{ristoski2016rdf2vec} presented \rtv which uses language modeling approaches for unsupervised feature extraction from sequences of words and adapts them to RDF graphs. 
After generating sequences by leveraging local information from graph substructures by random walks, \rtv learns latent numerical representations of entities in RDF graphs.
The algorithm has been extended in order to reduce the computational time and the biased regarded the random walking~\citep{cochez2017biased}.
More recently, \cite{cochez2017global} exploited the Global Vectors algorithm to compute embeddings from the co-occurrence matrix of entities and relations without generating the random walks.
In following research, the authors refer to their algorithm as \kgl.\footnote{\url{https://datalab.rwth-aachen.de/embedding/KGloVe/}}

\section{\ktvplain} \label{sec:core}

This study addresses the following research questions:

\begin{enumerate}
    \item Can we generate embeddings at a high rate while preserving accuracy?
    \item How can we test the distributional hypothesis of KGEs?
    \item Can we learn a scoring function for knowledge base completion which performs better than the standard one?
\end{enumerate}


Formally, let $t = (s,p,o)$ be a triple containing a subject, a predicate, and an object in a knowledge base $K$.
For any triple, $(s,p,o) \subseteq E \times R \times (E \cap L)$, where $E$ is the set of all entities, $R$ is the set of all relations, and $L$ is the set of all literals (i.e., string or numerical values).
A representation function $F$ defined as
\begin{equation}
F : (E \cap R \cap L) \rightarrow \mathbb{R}^d
\end{equation}
assigns a vector of dimensionality $d$ to an entity, a relation, or a literal.
However, some approaches consider only the vector representations of entities or subjects (i.e, $\lbrace s \in E : \exists (s, p, o) \in K \rbrace$).
For instance, in approaches based on Tensor Factorization, given a relation, its subjects and objects are processed and transformed into sparse matrices; all the matrices are then combined into a tensor whose depth is the number of relations.
For the final embedding, current approaches rely on dimensionality reduction to decrease the overall complexity~\citep{nickel2014reducing,TransA/jia2015locally,TransR/lin2015learning}.
The reduction is performed through an embedding map $\Phi : \mathbb{R}^d \rightarrow \mathbb{R}^k$, which is a homomorphism that maps the initial vector space into a smaller, reduced space.
The positive value $k < d$ is called the \textit{rank} of the embedding.
Note that each dimension of the reduced common space does not necessarily have an explicit connection with a particular relation.
Dimensionality reduction methods include Principal Component Analysis techniques~\citep{nickel2014reducing} and generative statistical models such as Latent Dirichlet Allocation~\citep{Sspace/jurgens2010s,Gensim/rehurek_lrec}.

Existing KGE approaches based on the skip-gram model such as \rtv~\citep{ristoski2016rdf2vec} submit paths built using random walks to a \wtv algorithm.
Instead, we preprocess the input knowledge base by converting each triple into a small sentence of three words.
Our method is faster as it allows us to avoid the path generation step.
The generated text corpus is thus processed by the skip-gram model as follows.

\subsection{Adapting the skip-gram model}

We adapt the skip-gram model~\citep{mikolov2013distributed} to deal with our small sequences of length three.
In this work, we only consider URIs and discard literals, therefore we compute a vector for each element $u \in E \cap R$.
Considering a triple as a sequence of three URIs $T = \{u_s, u_p, u_o$\}, the aim is to maximize the average log probability
\begin{equation}
\frac{1}{3} \sum_{u \in T} \sum_{u' \in T \setminus u} \log p(u | u')
\end{equation}
which means, in other words, to adopt a context window of $2$, since the sequence size is always $|T|=3$.
The probability above is theoretically defined as:
\begin{equation}
p(u | u') = \frac{\exp ( {v^O_{u}}^{\top} v^I_{u'} )}{\sum_{x \in E \cap R} \exp ( {v^O_{x}}^{\top} v^I_{u'} )}
\end{equation}
where $v^I_x$ and $v^O_x$ are respectively the input and output vector representations of a URI $x$.
We imply a negative sampling of $5$, i.e. $5$ words are randomly selected to have an output of $0$ and consequently update the weights.

\subsection{Scoring functions}

\subsubsection{Scoring by analogy}

Several methods have been proposed to evaluate word embeddings.
The most common ones are based on analogies~\citep{mikolov2013linguistic,levy2014linguistic}, where word vectors are summed up together, e.g.:
\begin{equation}
v["queen"] \approx v["king"] + v["woman"] - v["man"]
\end{equation}
An analogy where the approximation above is satisfied within a certain threshold can thus predict hidden relationships among words, which in our environment means to predict new links among entities~\cite{ristoski2016rdf2vec}.
The analogy-based score function for a given triple $(\bar{s},\bar{p},\bar{o})$ is defined as follows.
\begin{equation}
score(\bar{s},\bar{p},\bar{o}) = \frac{1}{\left\vert \{ (s,\bar{p},o) \in K \} \right\vert} \sum_{(s,\bar{p},o) \in K} {
\begin{cases}
    1 & \text{if } \left\Vert v_{\bar{s}} + v_o - v_s - v_{\bar{o}} \right\Vert \leq \epsilon \\
    0 & \text{otherwise}
\end{cases}
} 
\end{equation}
where $\epsilon$ is an arbitrarily small positive value.
In words, given a predicate $\bar{p}$, we select all triples where it occurs.
For each triple, we compute the relation vector as the difference between the object and the subject vectors.
We then count a match whenever the vector sum of subject $\bar{s}$ and relation is close to object $\bar{o}$ within a radius $\epsilon$.
The score is equal to the rate of matches over the number of selected triples.

\subsubsection{Scoring by neural networks}

We evaluate the scoring function above against a neural network based on Long Short-Term Memories (LSTM).
The neural network takes a sequence of embeddings as input, namely $v_s, v_p, v_o$ for a triple $(s,p,o) \in K$.
A dense hidden layer of the same size of the embeddings is connected to a single output neuron with sigmoid activation, which returns a value between $0$ and $1$.
The negative triples are generated using two strategies, i.e. for each triple in the training set (1) randomly extract a relation and its two nodes or (2) corrupt the subject or the object.
We use the Adam optimizer and 100 epochs of training.

\begin{figure}[t]
    \sidecaption
    \includegraphics[scale=0.3]{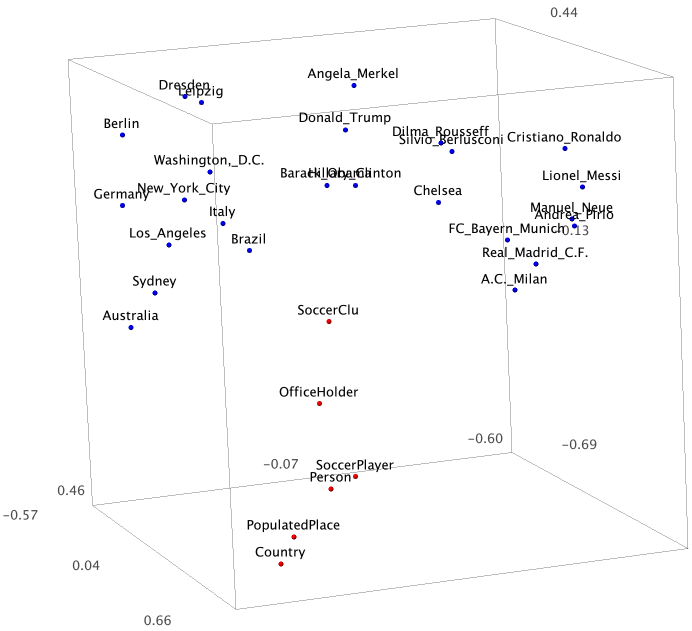}
    \caption{A selection of DBpedia resources along with their vectors in 3 dimensions obtained using Principal Component Analysis. Blue points are resources, whilst red points are classes. As can be seen, resources follow the distributional hypothesis.}
    \label{fig:dbpedia2vec}
\end{figure}

\subsection{Metrics}

As recently highlighted by several members of the ML and NLP communities, KGEs are rarely evaluated on downstream tasks different from link prediction (also known as knowledge base completion).
Achieving high performances on link prediction does not necessarily mean that the generated embeddings are good, since the inference task is often carried out in combination with an external algorithm such as a neural network or a scoring function.
The complexity is thus approach-dependent and distributed between the latent structure in the vector model and the parameters (if any) of the inference algorithm.
For instance, a translational model such as TransE~\cite{TransE/bordes2013translating} would likely feature very complex embeddings, since in most approaches the inference function is a simple addition.
On the other hand, we may find less structure in a tensor factorization model such as RESCAL~\cite{nickel2011three}, as the inference is performed by a feed-forward neural network which extrapolates the hidden semantics layer by layer.

\subsubsection{Neighbour Similarity Test}

In this paper, we introduce two metrics inspired by \textit{The Identity of Indiscernibles}~\citep{black1952identity} to gain insights over the distributional quality of the learned embeddings.
\begin{svgraybox}
The more characteristics two entities share, the more similar they are and so should be their vector representations.
\end{svgraybox}
Considering the set of characteristics $C_K(s)=\{(p_1,o_1),\dots,(p_m,o_m)\}$ of a subject $s$ in a triple, we can define a metric that expresses the similarity among two entities $e_1,e_2$ as the Jaccard index between their sets of characteristics $C_K(e_1)$ and $C_K(e_2)$.
Given a set of entities $\Tilde{E}$ and their $N$ nearest neighbours in the vector space, the overall \textit{Neighbour Similarity Test} (NST) metric is defined as:
\begin{equation} \label{eqn:nst}
    NST(\Tilde{E},N,K) = \frac{1}{N \vert \Tilde{E} \vert} \sum_{e \in \Tilde{E}} \sum_{j=1}^N \frac{\vert C_K(e) \cap C_K(n_j^{(e)}) \vert}{\vert C_K(e) \cup C_K(n_j^{(e)}) \vert}
\end{equation}
where $n_j^{(e)}$ is the $j$th nearest neighbour of $e$ in the vector space.

\subsubsection{Type and Category Test}

The second metric is the \textit{Type and Category Test} (TCT), based on the assumption that two entities which share types and categories should be close in the vector space.
This assumption is suggested by the human bias for which \texttt{rdf:type} and \texttt{dct:subject} would be predicates with a higher weight than the others.
Although this does not happen, we compute it for a mere sake of comparison with the NST metric.
The TCT formula is equal to Equation~\ref{eqn:nst} except for sets $C_K(e)$, which are replaced by sets of types and categories $TC_K(e)$.

\section{Evaluation} \label{sec:eval}


We implemented \ktv in Python 2.7 using the \textit{Gensim} and \textit{Keras} libraries with \textit{Theano} environment.
Source code, datasets, and vectors obtained are available online.\footnote{\url{http://github.com/AKSW/KG2Vec}}
All experiments were carried out on an Ubuntu 16.04 server with 128 GB RAM and 40 CPUs.

\begin{table}[ht]
\setlength{\tabcolsep}{2.5pt}
\centering
\caption{Details and runtimes for the generation of \ktv embeddings on two datasets.} \label{tab:data}
\begin{tabular}{lR{2cm}R{2cm}R{2cm}R{2cm}}
\toprule
\textbf{Dataset}  & \textbf{AKSW-bib}  & \textbf{DBpedia 2015-10}   & \multicolumn{2}{c}{\textbf{DBpedia 2016-04}} \\ \midrule
Number of triples & 3922                & 164,369,887               & 276,316,003 & 276,316,003 \\ 
Number of vectors & 954                & 14,921,691                 & 23,816,469 & 36,596,967 \\ \midrule
Dimensionality    & 10                & 300                         & 200        & 200 \\
Runtime (s)       & 2.2                & 18,332                     & 25,380     & 46,099 \\
Rate (triples/s)  & 1,604                & 8,966                    & 10,887     & 5,994 \\
\bottomrule
\end{tabular}
\end{table}

The dataset used in the experiments are described in Table~\ref{tab:data}.
The AKSW-bib dataset -- employed for the link prediction evaluation -- was created using information from people and projects on the AKSW.org website and bibliographical data from Bibsonomy.
We built a model on top of the English 2015-10 version of the DBpedia knowledge graph~\citep{dbpedia_jws_09}; Figure~\ref{fig:dbpedia2vec} shows a 3-dimensional plot of selected entities. 
For the English DBpedia 2016-04 dataset, we built two models.
In the first, we set a threshold to embed only the entities occurring at least 5 times in the dataset; we chose this setting to be aligned to the related works' models.
In the second model, all 36 million entities in DBpedia are associated a vector.
More insights about the first model can be found in the next two subsections, while the resource consumption for creating the second model can be seen in Figure~\ref{fig:cpumem}.

\subsection{Runtime}


In this study, we aim at generating embeddings at a high rate while preserving accuracy.
In Table~\ref{tab:data}, we already showed that our simple pipeline can achieve a rate of almost $11,000$ triples per second on a large dataset such as DBpedia 2016-04.
In Table~\ref{tab:runtime}, we compare \ktv with three other scalable approaches for embedding knowledge bases.
We selected the best settings of \rtv and \kgl according to their respective articles, since both algorithms had already been successfully evaluated on DBpedia~\citep{ristoski2016rdf2vec,cochez2017global}.
We also tried to compute \textit{fastText} embeddings on our machine, however we had to halt the process after three days.
As the goal of our investigation is efficiency, we discarded any other KGE approach that would have needed more than three days of computation to deliver the final model~\citep{cochez2017global}.

\rtv has shown to be the most expensive in terms of disk space consumed, as the created random walks amounted to $\sim$300 GB of text.
Moreover, we could not measure the runtime for the first phase of \kgl, i.e. the calculation of the Personalized PageRank values of DBpedia entities.
In fact, the authors used pre-computed entity ranks from \cite{Thalhammer2016} and the \kgl source code does not feature a PageRank algorithm.
We estimated the runtime comparing their hardware specs with ours.
Despite being unable to reproduce any experiments from the other three approaches, we managed to evaluate their embeddings by downloading the pretrained models\footnote{\url{http://data.dws.informatik.uni-mannheim.de/rdf2vec/}} and creating a \ktv embedding model of the same DBpedia dataset there employed.

\begin{table}[htbp]
\centering
\caption{Runtime comparison of the single phases. Those with (*) are estimated runtimes.}
\label{tab:runtime}
\begin{tabular}{@{}lll@{}}
\toprule
\textbf{Approach} & \textbf{Steps} & \textbf{Time} \\ \midrule
\multirow{2}{*}{\textbf{RDF2Vec}} & Random walks generation & 123 minutes \\
 & Word2Vec training & >96 hours (*) \\ \midrule
\multirow{3}{*}{\textbf{KGloVe}} & Personalized PageRank & N/A \\
 & Co-occurrence count matrix & \multirow{2}{*}{12 hours (*)} \\
 & GloVe training &  \\ \midrule
\multirow{2}{*}{\textbf{KG2Vec}} & Conversion to text & 5 minutes \\ 
 & Word2Vec training & 6 hours 58 minutes \\ \midrule 
\multirow{2}{*}{\textbf{fastText}} & Conversion to text & 5 minutes \\
 & fastText training & >72 hours (*) \\ \bottomrule
\end{tabular}
\end{table}

\vspace{-1cm}

\subsection{Preliminary results on link prediction}

For the link prediction task, we partition the dataset into training and test set with a ratio of 9:1.
In Table~\ref{tab:linkpred}, we show preliminary results between the different strategies on the AKSW-bib dataset using \ktv embeddings.
As can be seen, our LSTM-based scoring function significantly outperforms the analogy-based one in both settings.
According to the Hits@10 accuracy we obtained, corrupting triples to generate negative examples is the better strategy.
This first insight can foster new research on optimizing a scoring function for KGE approaches based on distributional semantics.

\begin{table}[htbp]
\setlength{\tabcolsep}{2.5pt}
\centering
\caption{Filtered Hits@10 values on link prediction on \textit{AKSW-bib} using different strategies.}
\label{tab:linkpred}
\begin{tabular}{lR{2cm}R{2cm}R{2cm}}
\toprule
      & Hits@1 & Hits@3 & Hits@10 \\ \midrule
LSTM + corrupted    & 3.84\% & 9.79\% & 19.23\% \\
LSTM + random       & 1.39\% & 4.89\% & 10.49\% \\
Analogy             & 0.00\% & 0.51\% & 3.82\% \\
\bottomrule
\end{tabular}
\end{table}


\vspace{-1cm}

\subsection{Distributional quality}

\begin{figure}[htbp]
    \begin{multicols}{2}
    \includegraphics[width=\linewidth]{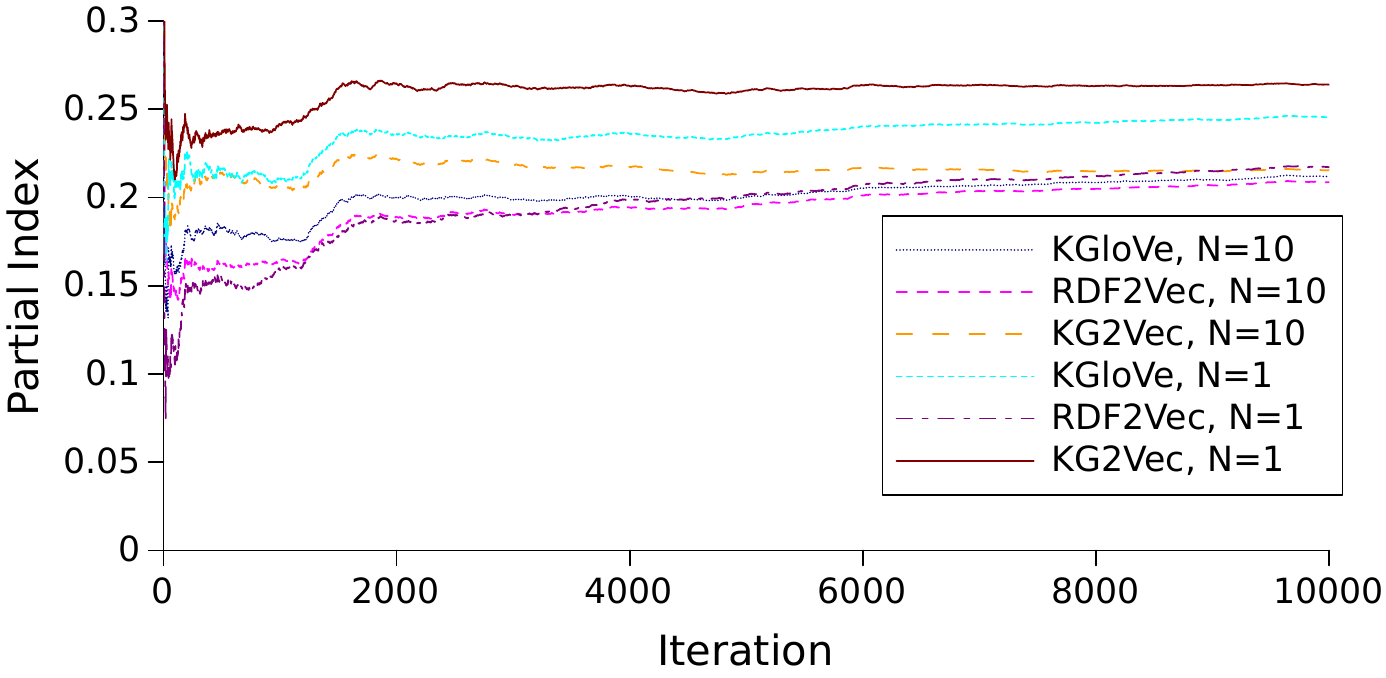}\par
	\caption{Partial TCT value on DBpedia 2016-04 for the top 10,000 entities.}
	\label{fig:tct10top}
    \includegraphics[width=\linewidth]{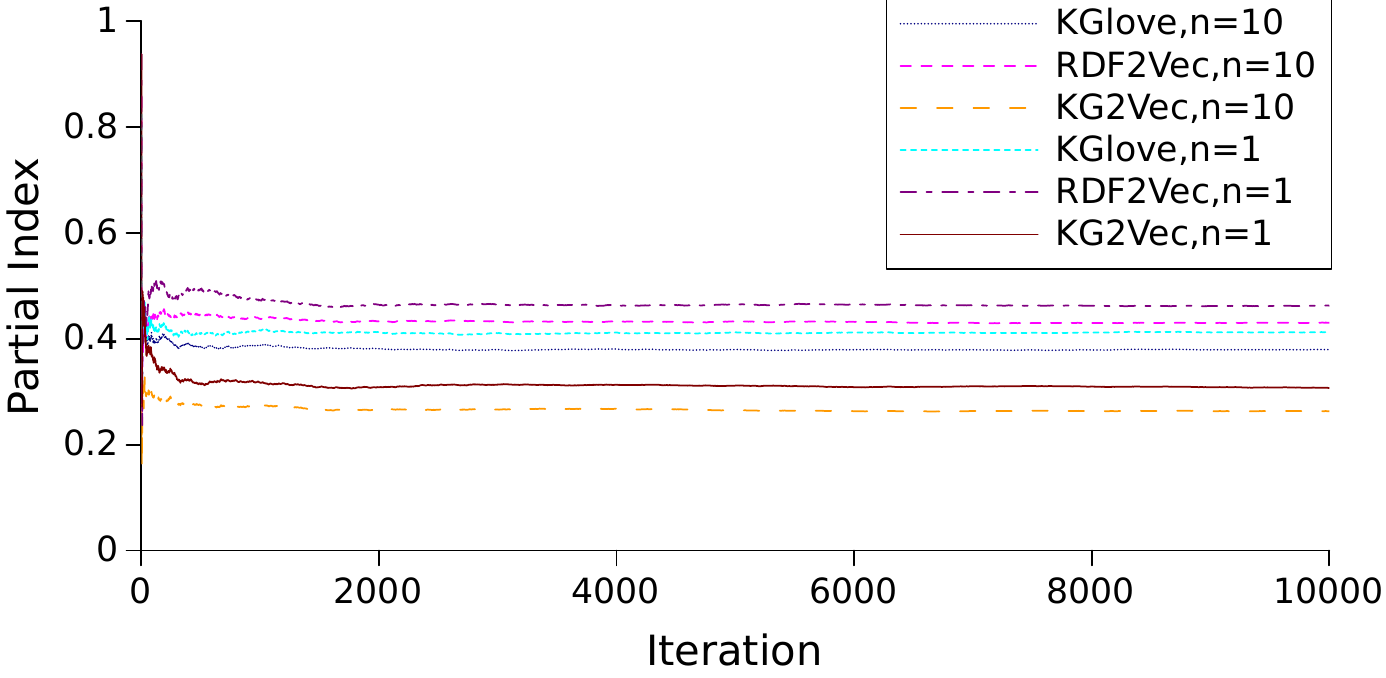}\par
	\caption{Partial TCT value on DBpedia 2016-04 for 10,000 random entities.}
	\label{fig:tct10rand}
    \end{multicols}
\end{figure}

\begin{figure}[htbp]
    \begin{multicols}{2}
    \includegraphics[width=\linewidth]{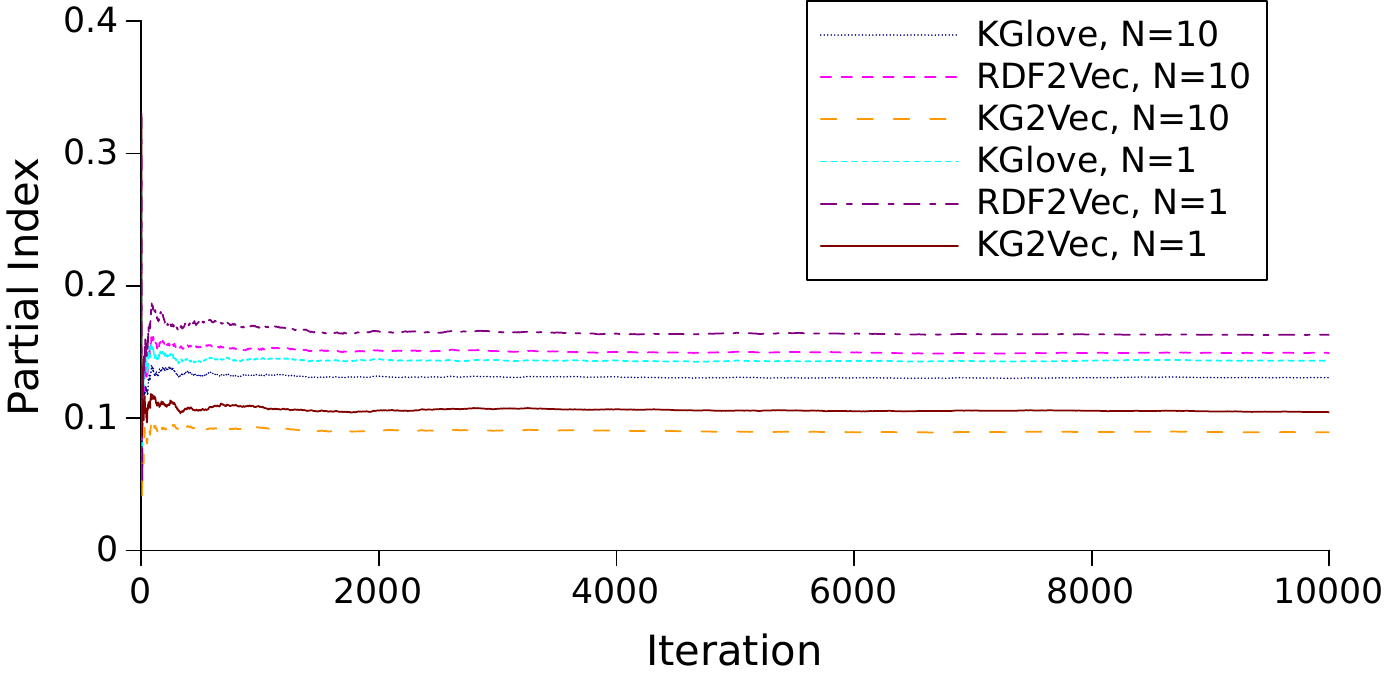}\par
	\caption{Partial NST value on DBpedia 2016-04 for 10,000 random entities.}
	\label{fig:nst10rand}
    \includegraphics[width=\linewidth]{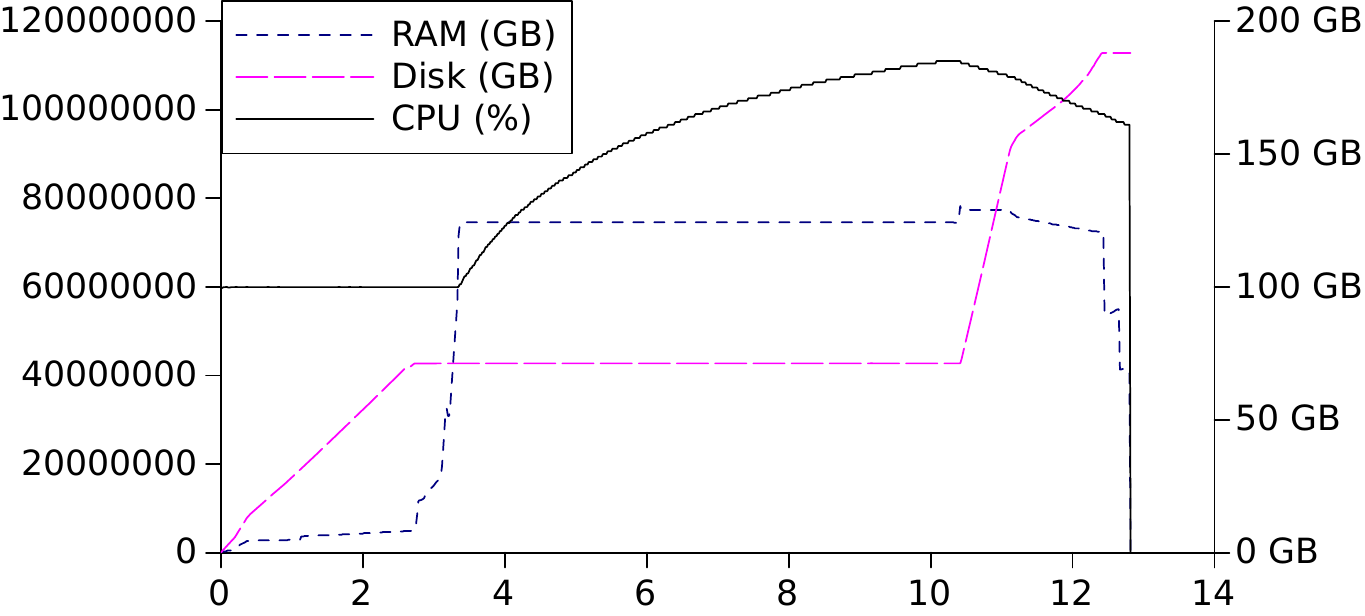}\par
	\caption{CPU, Memory, and disk consumption for \ktv on the larger model of DBpedia 2016-04.}
	\label{fig:cpumem}
    \end{multicols}	
\end{figure}

Computing the NST and TCT distributional quality metrics on the entire DBpedia dataset is time-demanding, since for each entity, the model and the graph need to be queried for the $N$ nearest neighbours and their respective sets.
However, we approximate the final value by tracing the partial values of NST and TCT over time.
In other words, at each iteration $i$, we compute the metrics over $\Tilde{E}_i = \{e_1, \dots, e_i\}$.
Figure~\ref{fig:tct10top} shows the partial TCT value on the most important 10,000 entities for $N=\{1,10\}$ according to the ranks computed by \cite{Thalhammer2016}.
Here, \ktv maintains a higher index than the other two approaches, despite these are steadily increasing after the $\sim2,000$th entity.
We interpret the lower TCT for the top $2,000$ entities as noise produced by the fact that these nodes are hyperconnected to the rest of the graph, therefore it is hard for them to remain close to their type peers.
In Figures~\ref{fig:tct10rand} and \ref{fig:nst10rand}, the TCT and NST metrics respectively are computed on 10,000 random entities.
In both cases, the values for the two settings of all approaches stabilize after around $1,000$ entities, however we clearly see that \rtv embeddings achieve the highest distributional quality by type and category.
The higher number of occurrences per entity in the huge corpus of random walks in \rtv might be the reason of this result for rarer entities.


\begin{figure}[htbp]
    \sidecaption
    \includegraphics[width=0.5\linewidth]{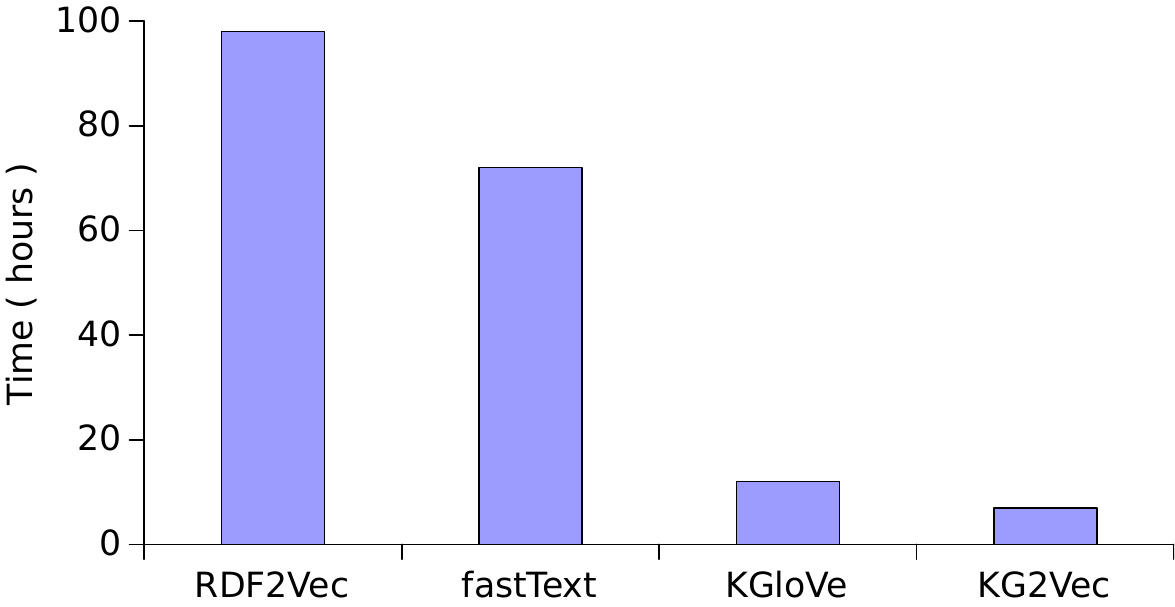}
	\caption{We show the comparison of the run-times for all four approaches. Note that since we do not know how long the PageRank computation takes, we reported the estimated runtime for the plain version of \kgl.}
	\label{fig:runtime}
\end{figure}

In Figure~\ref{fig:cpumem}, we show the CPU, Memory, and disk consumption for \ktv on the larger model of DBpedia 2016-04.
All three subphases of the algorithm are visible in the plot.
For 2.7 hours, tokens are counted; then, the learning proceeds for 7.7 hours; finally in the last 2.3 hours, the model is saved.



    

\section{Conclusion and Future Work} \label{sec:conclusion}

We presented a fast approach for generating KGEs dubbed \ktv.
We conclude that the skip-gram model, if trained directly on triples as small sentences of length three, significantly gains in runtime while preserving a decent vector quality.
Moreover, the \ktv embeddings have shown higher distributional quality for the most important entities in the graph according to PageRank.
As a future work, we plan to extend the link prediction evaluation to other benchmarks by using analogies and our LSTM-based scoring function over the embedding models of the approaches here compared.

%
%

\FloatBarrier
\bibliographystyle{spbasic}
{
\FloatBarrier
\bibliography{biblio}
}

\end{document}